\documentclass[journal,article,submit,pdftex,moreauthors]{mdpi} 
\firstpage{1} 
\pubvolume{1}
\issuenum{1}
\articlenumber{0}
\pubyear{2022}
\copyrightyear{2022}
\datereceived{} 
\dateaccepted{} 
\datepublished{}
\hreflink{https://doi.org/} % If needed use \linebreak
\Title{$ \text{T}^3 $OMVP: A Transformer-based Time and Team Reinforcement Learning Scheme for Observation-constrained Multi-Vehicle Pursuit in Urban Area}

\TitleCitation{$ \text{T}^3 $OMVP: A Transformer-based Time and Team Reinforcement Learning Scheme for Observation-constrained Multi-Vehicle Pursuit in Urban Area}

\Author{Zheng Yuan %Please carefully check the accuracy of names and affiliations. 
 $^{1}$\orcidA{}, Tianhao Wu $^{1}$, Qinwen Wang $^{1}$, Yiying Yang $^{1}$, Lei Li $^{1}$ and Lin Zhang $^{1}$*}
 
\AuthorNames{Zheng Yuan, Tianhao Wu, Qinwen Wang, Yiying Yang, Lei Li, Lin Zhang}

\AuthorCitation{Yuan, Z.; Wu, T.; Wang, Q.; Yang Y.; Li, L.; Zhang, L.}
\address[1]{%
School of Artificial Intelligence, Beijing University of Posts and Telecommunications, 10 Xitucheng Road, \linebreak~Haidian Distinct, Beijing 100876, China;   
yuanzheng@bupt.edu.cn (Z.Y.); wu.tianhao@bupt.edu.cn (T.W.);
wangqinwen@bupt.edu.cn (Q.W.); yyying@bupt.edu.cn (Y.Y.); leili@bupt.edu.cn (L.L.); zhanglin@bupt.edu.cn (L.Z.)
}

\corres{\hangafter=1 \hangindent=1.05em \hspace{-0.82em}Correspondence: zhanglin@bupt.edu.cn}

\abstract{Smart Internet of Vehicles (IoVs) combined with Artificial Intelligence (AI) will contribute to vehicle decision-making in the Intelligent Transportation System (ITS). 
Multi-Vehicle Pursuit games (MVP), a multi-vehicle cooperative ability to capture mobile targets, is becoming a hot research topic gradually. 
Although there are some achievements in the field of MVP in the open space environment, the urban area brings complicated road structures and restricted moving spaces as challenges to the resolution of MVP games. We define an Observation-constrained MVP (OMVP) problem in this paper and propose a Transformer-based Time and Team Reinforcement Learning scheme ($ \text{T}^3 $OMVP) to address the problem.
First, a new multi-vehicle pursuit model is constructed based on decentralized partially observed Markov decision processes (Dec-POMDP) to instantiate this problem.
Second, by introducing and modifying the transformer-based observation sequence, QMIX is redefined to adapt to the complicated road structure, restricted moving spaces and constrained observations, so as to control vehicles to pursue the target combining the vehicle's observations. 
Third, a multi-intersection urban environment is built to verify the proposed scheme. 
Extensive experimental results demonstrate that the proposed $ \text{T}^3 $OMVP scheme achieves significant improvements relative to state-of-the-art QMIX approaches by 9.66\%\textasciitilde106.25\%. Code is available at \url{https://github.com/pipihaiziguai/T3OMVP}.} 


\begin{document}

%%%%%%%%%%%%%%%%%%%%%%%%%%%%%%%%%%%%%%%%%%
% \setcounter{section}{-1} %% Remove this when starting to work on the template.
%\section{How to Use this Template}

%The template details the sections that can be used in a manuscript. Note that the order and names of article sections may differ from the requirements of the journal (e.g., the positioning of the Materials and Methods section). Please check the instructions on the authors' page of the journal to verify the correct order and names. For any questions, please contact the editorial office of the journal or support@mdpi.com. For LaTeX-related questions please contact latex@mdpi.com.%\endnote{This is an endnote.} % To use endnotes, please un-comment \printendnotes below (before References). Only journal Laws uses \footnote.

% The order of the section titles is: Introduction, Materials and Methods, Results, Discussion, Conclusions for these journals: aerospace,algorithms,antibodies,antioxidants,atmosphere,axioms,biomedicines,carbon,crystals,designs,diagnostics,environments,fermentation,fluids,forests,fractalfract,informatics,information,inventions,jfmk,jrfm,lubricants,neonatalscreening,neuroglia,particles,pharmaceutics,polymers,processes,technologies,viruses,vision

\section{Introduction}

The Internet of Vehicles (IoVs) is a typical application of Internet of Things (IoT) technologies in the Intelligent Transportation System (ITS) \cite{A survey,Driverless transportation system,internet of vehicles}. With the Vehicle-to-Vehicle (V2V) and Vehicle-to-Infrastructure (V2I) communications, IoVs combined with Artificial Intelligence (AI) can improve the decision-making of vehicles and the efficiency of ITS \cite{IoV decision1,wu2020cooperative}. The Multi-Vehicle Pursuit (MVP) game describes a multi-vehicle cooperative ability to capture mobile targets, represented by the New York City Police Department guideline on the pursuit of suspicious vehicles \cite{NYCPD}. The MVP game is becoming a hot topic in the IoVs supported by AI research.

There are mainly two ways to solve the MVP game, one is the game theory, and the other is cooperative multi-agent reinforcement learning (MARL). With respect to game theory, Eloy et al. proposed a team cooperative optimal solution of the border-defense differential game \cite{garcia2020multiple}. Huang et al. proposed a decentralized control scheme based on the Voronoi partition of the game domain \cite{huang2011guaranteed}. However, it becomes challenging for game theory solutions to define a suitable objective function when the problem becomes more complex, such as the increased number of agents, restricted movement, and complicated environment.

\begin{figure}
  \includegraphics[width=\linewidth]{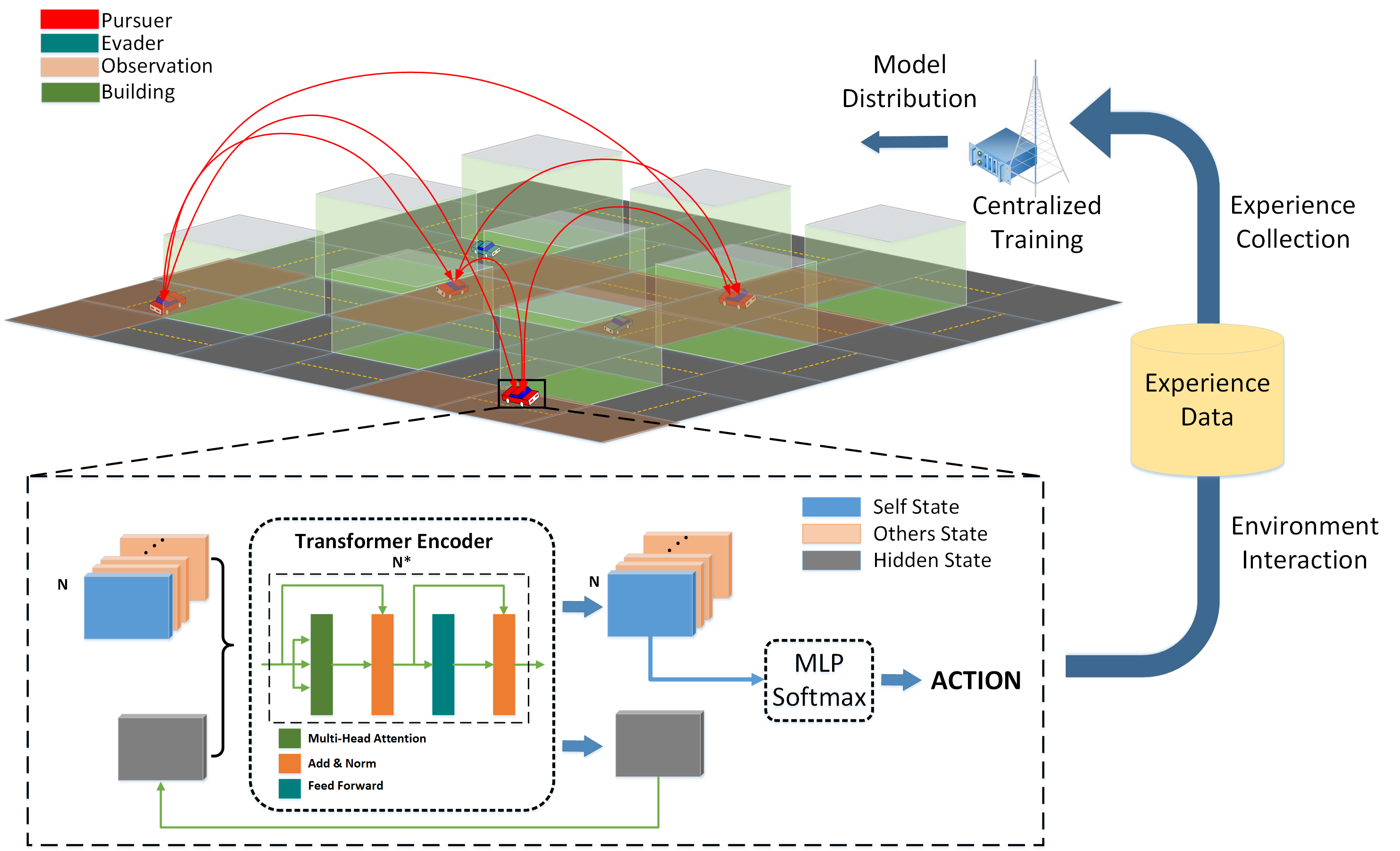}
  \caption{An overview of $ \text{T}^3 $OMVP for the problem of multi-vehicle pursuit in the multi-intersection urban area.}
  \label{fig:General} %% label for entire figure
\end{figure}

With respect to cooperative MARL research for the MVP game, the multi-agent system is modeled using Markov decision processes (MDP) \cite{RL2}, and a neural network can be used to approximate the complex objective function \cite{qu2021adaptive}. Cristino et al. used the Twin Delayed Deep Deterministic Policy Gradient (TD3) to demonstrate a real-world pursuit-evasion in open environment with boundaries \cite{de2021decentralized}. Timothy used the Deep Deterministic Policy Gradient (DDPG) with omnidirectional agents \cite{lillicrap2015continuous}. Thomy et al. proposed a Resilient Adversarial value Decomposition with Antagonist-Ratios method and verified the method on the predator-prey scenario \cite{resilient}. Jiang et al. proposed a vehicular end-edge-cloud computing framework to implement vertical and horizontal cooperation among vehicles \cite{jiang2020multi}. Peng et al. proposed a Coordinated Policy Optimization to facilitate the cooperation of agents at both local and global levels \cite{peng2021learning}. However, the above studies in cooperative MARL are not verified in urban environments with complicated road structures, restricted moving spaces and especially constrained observations due to architectural obstructions. 

This paper introduces the transformer block to process the observation-entities of all agents inspired by \cite{updet}. QMIX \cite{rashid2018qmix} is used as the baseline to control the pursuing vehicles, which is a state-of-the- art MARL algorithm that was successfully applied to other domains. Furthermore, this paper modifies the predator-prey scenario in \cite{resilient} by adding intersections to the original open space to construct a multi-vehicle pursuing urban area for instantiating the MVP game. The proposed Transformer-based Time and Team Reinforcement Learning scheme ($ \text{T}^3 $OMVP) is presented in Figure \ref{fig:General}.

The main contributions of this paper are:

1.	The MVP game in the urban environment is defined as the Observation-constrained Multi-Vehicle Pursuit (OMVP) problem with the occlusion of buildings to each pursuing vehicle.

2.  The $ \text{T}^3 $OMVP scheme is proposed to address the OMVP problem using partially observable Markov decision process and reinforcement learning. The $ \text{T}^3 $OMVP introduces the transformer block to deal with the observation sequences to realize the multi-vehicle cooperation without policy decoupling strategy. 

3.	A novel low-cost and light-weight MVP simulation environment is constructed to verify the performance of $ \text{T}^3 $OMVP scheme. Compared with large-scale game environments, this simulation environment can improve the training speed of reinforcement learning. All source codes are provided on GitHub (\url{https://github.com/pipihaiziguai/T3OMVP}).

The details of the MVP problem and its relationship with Dec-POMDP are presented in Section \ref{sec:MVP}. The OMVP problem and specific implementation details of the $ \text{T}^3 $OMVP scheme are presented in Section \ref{sec:OMVP}. The experimental results and discussions are provided in Section \ref{sec:result}. Section \ref{sec:conclusion} concludes this paper.

%%%%%%%%%%%%%%%%%%%%%%%%%%%%%%%%%%%%%%%%%%

\section{Multi-Vehicle Pursuit}
\label{sec:MVP}

\subsection{Problem Statement}
This paper mainly describes a multi-intersection urban area, as shown in Figure \ref{fig:General}. In a $ W*W $ grid, there are $ N $ learning red pursuing vehicles and $ \frac{N}{2} $ randomly moving blue evading vehicles. The pursuing vehicle has an observation field of $ M*M $, but the shape of its observation field is a cross or a straight line corresponding to the scenarios of an intersection or a straight road due to the influence of obstacles. According to the description in \cite{NYCPD}, each pursuing vehicle will share its observation and position with other pursuing vehicles. The shared observation is the position of the evading vehicle in field of vision. In addition, each vehicle also has an observation field showing the position of obstacles in its $ M*M $ field. Therefore, the $ state $ of each pursuing vehicle includes its own $ M*M $ observation and those shared by other vehicles. The $ global$ $ state $ is a $ W*W $ observation. The $ reward $ of each pursuing vehicle is calculated by the normalized function $ g = \frac{1}{n} $. If there are $ n \in \{1,\ldots,N\}$ pursuing vehicles capturing the same evading vehicle, the $ reward $ of each corresponding pursuing vehicle is $ \frac{1}{n} $. The $ global $ $ reward $ is the sum of all pursuing vehicles rewards, and an evading vehicle is captured with the $ global$ $ reward $ of +1.

\subsection{Dec-POMDP}
On the scenario of urban area, each vehicle can be modeled as an agent which can only have partial observation of the environment. As such, the cooperative multi-agent MVP task characterized by communications between agents can be formulated as partially observable Markov game  $G=\langle S, K, A, P, r, O, Z, n, \gamma\rangle$. The global state of the environment is denoted by $s \in S$. At each step $t$, each agent $ k \in \mathbf{K}\equiv\{1, \ldots, N\}$ chooses an action $a^{k} \in A$, forming a joint action $\mathbf{a} \in \mathbf{A} \equiv A^{n} $. Thus the state transition function $P\left(s^{\prime} \mid s, \mathbf{a}\right): S \times \mathbf{A} \times S \rightarrow[0,1]$ represents a transition in the environment. A global reward function $r\left( s, \mathbf{a}\right): S \times \mathbf{A} \rightarrow R$ is necessary in MARL to estimate a  policy, this paper shares the same reward function among agents. Meanwhile, a partially observable scenario is considered in which each agent draws individual observation $ o \in O$ according to observation function $Z\left( s, k\right): S \times K \rightarrow O$. Each agent has an action-observation history $\tau^{k} \in T \equiv(O \times K)^{*}$ on which it conditions a stochastic policy $\pi^{k}\left(a^{k} \mid \tau^{k}\right): T \times A \rightarrow [0,1]$. The joint policy $\pi$ has a joint action-value function $Q^{\pi}\left(s_{t}, \mathbf{a}_{t}\right)=\mathbb{E}_{s_{t+1: \infty}, \mathbf{a}_{t+1: \infty}}\left[R_{t} \mid s_{t}, \mathbf{a}_{t}\right]$, where $R_{t}=\sum_{i=0}^{\infty} \gamma^{i} r_{t+i}$ is a discounted reward and $\gamma \in [0,1)$ is the discount factor.

\section{$ \text{T}^3 $OMVP scheme}
\label{sec:OMVP}
The OMVP problem can be formulated by cooperative MARL, each pursuing vehicle can be modeled as an agent with state, action, and reward.

\subsection{State, Action, and Reward}
According to the assumptions of MVP problem, each pursuing vehicle can obtain the observations of other pursuing vehicles through V2V or V2I communication. At time $t$, $ p_{\text{pur,k}}^t = (i_k^t,j_k^t) $ represents the position of the $ k^{\text{th}} $ pursuer, $ p_{\text{eva,m}}^t = (i_m^t,j_m^t) $ represents the position of the $ m^{\text{th}} $ evader, and $ p_{\text{obs,n}} = (i_n,j_n) $ represents the position of the $ n^{\text{th}} $ obstacle area. Therefore, $ P_{\text{pur}}^t=(p_{\text{pur,1}}^t, \ldots,p_{\text{pur,N}}^t) $ means the positions of all pursuers, $ P_{\text{eva}}^t=(p_{\text{eva,1}}^t, \ldots,p_{\text{eva,N/2}}^t) $ means the positions of all evaders, and $ P_{\text{obs}} = (p_{\text{obs,1}}, \ldots) $ means the positions of all obstacles areas. It is defined that the observation of the $ k^{\text{th}} $ pursuer at time $ t $ is divided into two parts: evading target observation and obstacle observation, which are represented by $ E_k^t $ and $ B_k^t $, respectively. Both $ E_k^t $ and $ B_k^t $ are represented by an $ M*M $ matrix, where $ E_k^t $ is limited by urban roads and $ B_k^t $ is not limited. $ e_{\text{k,i,j}}^t $ and $ b_{\text{k,i,j}}^t $ are the elements in row $ i $ and column $ j $ in $ E_k^t $ and $ B_k^t $, respectively. Define $ \emph{o}_k^t \in E_k^t $ as the observable area of the $ k^{\text{th}} $ pursuer, whose shape is a cross or a straight line. For the OMVP problem in grid environment, let
\begin{equation}\label{eqexpmuts:1}
\begin{aligned}[b]
e_{\text{k,i,j}}^t & = \begin{cases} 1 & if \thinspace (i,j) \in \emph{o}_k^t \thinspace  and \thinspace (i,j) \in P_{\text{eva}}^t  \\ 0 &  others \end{cases}  \\
b_{\text{k,i,j}}^t & = \begin{cases} 1 & if \thinspace (i,j) \in P_{\text{obs}}  \\ 0 &  others \end{cases} .
\end{aligned}
\end{equation}
In order to achieve vehicle cooperative pursuit, the $ state $ of each pursuing vehicle should include the observations of all other vehicles. $s_i^t = (E_i^t, B_i^t, E_{\text{other}}^t)$ represents the $ state $ of $ i^{\text{th}} $ pursuer at the time $ t $, including the  evading vehicle observation, the obstacle observation of $ i^{\text{th}} $ pursuer, and the evading vehicle observation of the other pursuers.

Since the urban area is built in the grid world, the action space of pursuing vehicles can be divided into five parts, including moving forward, moving backward, turning left at the intersection, turning right at the intersection, and stopping. The $ reward $ and $ global$ $ reward $ in OMVP are the same as in MVP.

\subsection{Centralized training with decentralized execution}
This paper considers that extra state information is available and vehicles can communicate freely.
Centralized training with decentralized execution (CTDE) is a standard paradigm in MARL. In training process, the centralized value function which conditions on global state and the joint actions is obtained. Meanwhile, each agent utilizes individual action-observation histories to learn its individual value function, which is updated by a centralized gradient provided by the centralized value function. In executing process, the learnt policy for each agent conditioning on individual value function can be executed in a decentralized way. State-of-the-art MARL algorithms, like VDN or QMIX, adopt this architecture.

\subsection{Observation-constrained Multi-Vehicle Pursuit}
OMVP can be modeled as a Dec-POMDP. This paper uses QMIX to deal with the multi-agent credit assignment in MVP. At the same time, in order to deal with the problem that the observation area is affected by the complex environment, the transformer is used to process the time observations and team observations.

\subsubsection{Monotonic Value Function Factorisation}
At present, cooperative MARL often uses CTDE for training but requires a centralized action-value function $ Q_{\text{total}} $ that conditions on the global state and the joint action. $ Q_{\text{total}} $ can usually be approximated using a neural network by  $ \hat{Q}_{\text{total}} $, which is factorized into approximate individual $ \hat{Q}_i $ for each agent $ i $ in order to update $ \hat{{\pi}}^i $ according to eq.\ref{eqexpmuts:2}. VDN is the simplest method to approximate the individual action-value function. It formulates $ \hat{Q}_{\text{total}} $ as a sum of individual action-value functions $ \hat{Q}_i(\tau_t^i,a_t^i) $, one for each agent $ i $, which condition only on individual action-observation histories:
\begin{equation}\label{eqexpmuts:2}
\begin{aligned}
\hat{Q}_{\text{total}}(\boldsymbol{\tau}_t, \mathbf{a}_t) = \sum_{i=1}^{N} \hat{Q}_i(\tau_t^i,a_t^i)
\end{aligned}
\end{equation}
Since the full factorisation of VDN is not necessary to extract decentralised policies that are fully consistent with their centralised counterpart, QMIX uses a mixing network to represent the $ \hat{Q}_{\text{total}} $. The mixing network is a feed-forward neural network that takes the agent network outputs as input and mixes them monotonically. The weights of the mixing network are restricted to be non-negative to enforce the monotonicity constraint of eq.\ref{eqexpmuts:3}.
\begin{equation}\label{eqexpmuts:3}
\begin{aligned}
\frac{\partial \hat{Q}_{\text{total}}}{\partial \hat{Q}_i} \geq 0, \forall i \in \mathbf{K}
\end{aligned}
\end{equation}

\subsubsection{Observations Sequence}
For the OMVP problem, although the state acquired by the pursuing vehicle is incomplete, more historical observations endow pursuing vehicles with better decisions of the next action. As such, the MVP problem becomes a Dec-POMDP. A coordinated method of time observations and team observations named $ TT-Observations $ is used to process the joint observation of all pursuing vehicles, thus solving the problem of constrained observations limited by the complex urban environment. 
\begin{figure}
  \includegraphics[width=\linewidth]{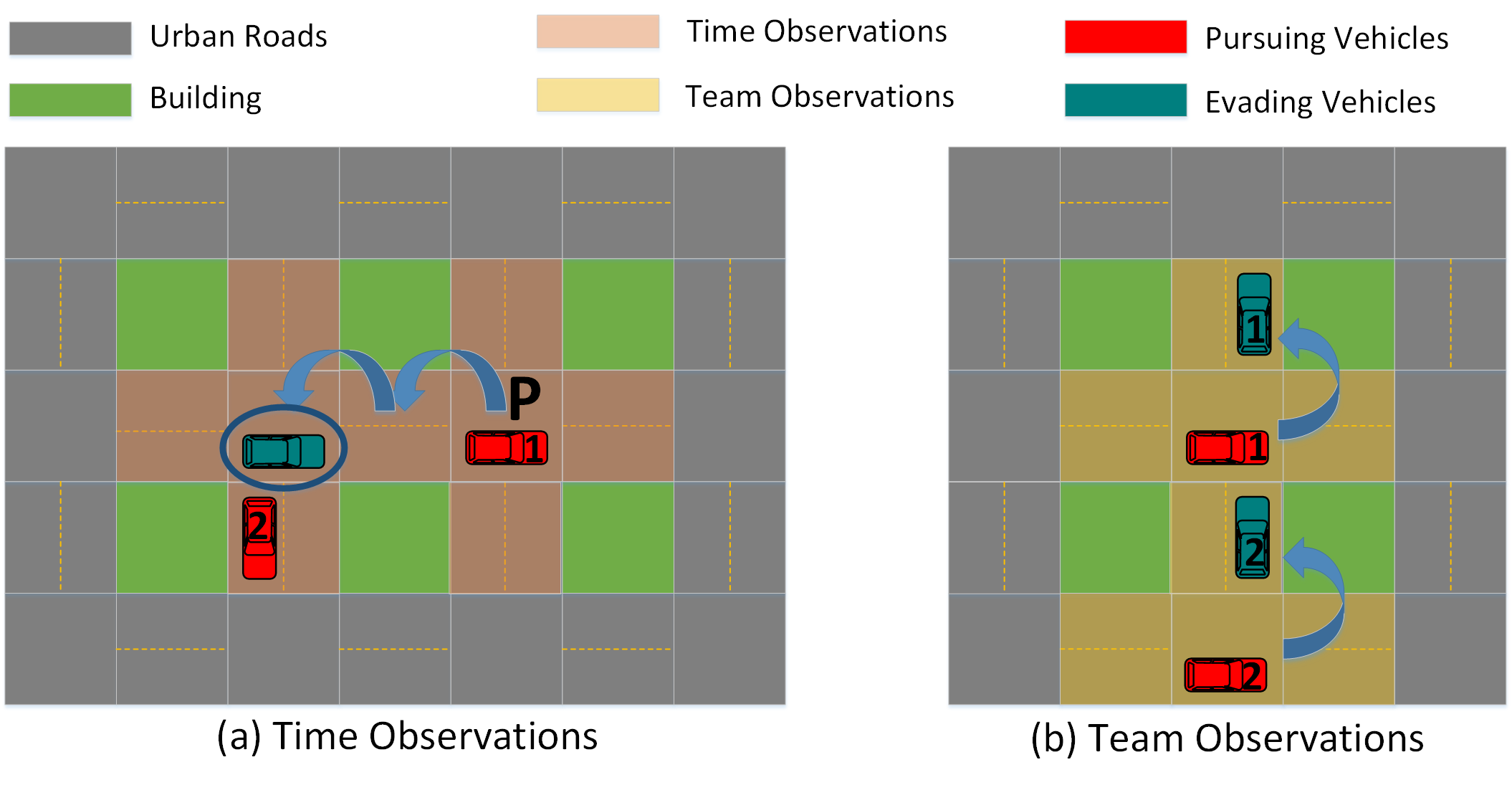}
  \caption{(a) Time observations are used to backtrack historical paths. (b) Team observations are used to assign team goals.}
  \label{fig:2}
\end{figure}
\paragraph{Time Observations}
Since pursuing vehicles can save part of the road information in the historical observation, a more extensive observation range can be obtained by combining multiple historical observations. Therefore, when other pursuing vehicles find the position of the target vehicle in the historically observed road, the pursuing vehicle can trace the source, thereby improving the pursuing efficiency. For example, in Figure \ref{fig:2} (a), the orange area is the historical time observations of the red pursuing vehicle No. 1. When it arrives at the position $ P $, an evading vehicle appears two blocks behind it. Due to the limitation of its field of view, pursuing vehicle No. 1 cannot observe the evading vehicle. Although the red pursuing vehicle No.2 can observe the evading vehicle, it still cannot pursue the evading vehicle after turning around, due to the evading vehicle can move to the right. Since the red pursuing vehicle No. 1 can drive in the opposite direction, it can trace the source based on historical time observations to capture the evading vehicle. Furthermore, the previously observed information is added to the input of the transformer block to process time observations using the self-attention mechanism.
\paragraph{Team Observations}
A more comprehensive observation range can also be constructed through integrating the observations of all pursuing vehicles. In IoVs, vehicles can communicate through V2V or V2I, so the team's observation sequences can be used to deal with complicated roads in urban environments. When multiple pursuing vehicles observe a same target, it is possible to avoid all pursuing vehicles aiming at the same target by focusing on the observation sequence of all vehicles, so that the target can be dispersed and the pursuing efficiency can be improved. For example, in Figure \ref{fig:2} (b), the yellow area is observations of all the red pursuing vehicles. When two pursuing vehicles observe the blue evading vehicle No. 2 at the same time, by processing all observation sequences, it is possible to avoid two red vehicles pursuing the blue evading vehicle No. 2 at the same time, and pursue the most suitable vehicle separately, thereby improving the pursuing efficiency. Furthermore, the joint observation of all pursuing vehicles is used as the input of the transformer block to process team observations using the self-attention mechanism.

\paragraph{$ TT-Observations $}
There are two ways to combine time observations and team observations. One is that the team observations are the main body, and the time observations are used as an additional input. The other is that the time observations are the main body, and the team observations are used as an additional input. In recurrent neural network, time information is usually transmitted and encoded through the hidden layer. Using the transformer structure based on team observations, the self-attention mechanism can be fully utilized to realize the cooperative control of the team. If the transformer structure based on time observations is used, the self-attention will only be placed on a single pursuing vehicle and cannot realize the cooperative control of the team. Therefore, $ \text{T}^3 $OMVP adopts the transformer structure based on team observations and uses a hidden layer to store time observations as additional input.

\subsubsection{Transformer-based $ TT-Observations $}
This paper uses the method $ TT-Observations $ that combines time observations and team observations and uses the transformer block to calculate the Q-value of each agent through the self-attention mechanism based on UPDeT.

\paragraph{Self-Attention Mechanism}
Vaswani et al. first used the self-attention mechanism in the transformer block \cite{attention}. In the self-attention, each input embedding vector $ \emph{X} $ has three different vectors, $ Q $, $ K $, and $ V $, representing query, key, and value, respectively. $ Q $, $ K $, $ V $ are obtained by multiplying three different weight matrices $ W_q $, $ W_k $, $ W_v $ with the embedding vector $ \emph{X} $, and the dimensions of the three weight matrices are the same. The calculation formula of self-attention output is as follows:
\begin{equation}\label{eqexpmuts:4}
\begin{aligned}
\text { Attention }(Q, K, V)=\operatorname{softmax}\left(\frac{Q K^{T}}{\sqrt{d_{k}}}\right) V.
\end{aligned}
\end{equation}
Here, $d_{k}$ is equal to the number of columns of the $ Q $ and $ K $ matrices to prevent the inner product from being too large.

\paragraph{Transformer in Time and Team Observations}
This paper assumes that the observation can be transmitted among the pursuing vehicles through communication. At time t, the observation set of all pursuing vehicles is $ {O}^t $, which contains independent observations of $ N $ agents $ \{o_1^t,o_2^t,\ldots,o_N^t\} $. The independent observations of all pursuing vehicles can be encoded into embedding vectors of the same length through the embedding layer, so the embedding vector of the entire team observations can be obtained as follows:
\begin{equation}\label{eqexpmuts:5}
\begin{aligned}
E_t = \{ \text{Emb}(o_1^t), \text{Emb}(o_2^t),\ldots, \text{Emb}(o_N^t) \}.
\end{aligned}
\end{equation}
Here, $ \text{Emb} $ represents the embedding layer in the neural network. Unlike UPDeT, this paper uses the joint observation of all pursuing vehicles as the input of the transformer block to realize the overall decision-making control of the pursuing vehicles. Similar to UPDeT, this paper uses a hidden state $ h_{t-1} $ to record the historical observation state to achieve historical backtracking. The observed embedding vector and the hidden state are combined at time $ t-1 $ as the input of the transformer block to obtain $ \emph{X}_1 $, and $ \emph{X}_l $ $( l \in \{1,2,\ldots,L\} )$ is the input of the transformer block of the $ l_{th} $ layer. The whole calculation process is as follows:
\begin{equation}\label{eqexpmuts:6}
\begin{aligned}
\emph{X}_1 & = \{ E_t, h_{t-1} \} \\
\{Q_l, K_l, V_l\} & = \emph{X}_l\{W_q, W_k, W_v\} \\
\emph{X}_{l+1} & = \text { Attention }(Q_l, K_l, V_l).
\end{aligned}
\end{equation}
Where $ W_q $, $ W_k $, $ W_v $ represent three matrices of the same shape for calculating $ Q $, $ K $, and $ V $, respectively. The $ \text{T}^3 $OMVP scheme further uses the output of the last self-attention layer as the input of the linear layer $ \text{LN} $ that calculates Q-value:
\begin{equation}\label{eqexpmuts:7}
\begin{aligned}
Q^t(E_t, h_{t-1}, \mathbf{a}) = \text{LN}(\emph{X}_{l+1}, \mathbf{a}) = \{Q_1^1,Q_2^t,\ldots,Q_N^t\}.
\end{aligned}
\end{equation}
Here, $ Q_i^t(i \in \{1,2,\ldots,N\}) $ represents the Q-value of each agent at time $ t $, and $ Q^t $ represents the set of Q-values of all agents. After obtaining the Q-values of all agents, the global Q-value can be calculated by credit assignment function:
\begin{equation}\label{eqexpmuts:8}
\begin{aligned}
Q_{\text{total}}^t(\boldsymbol{\tau}_t, \mathbf{a}_t) = \text{F}(Q_1^1,Q_2^t,\ldots,Q_N^t)
\end{aligned}
\end{equation}
where $ \text{F} $ is the credit assignment function. In this paper, QMIX is used to calculate the global Q-value. In QMIX, F is a mixed linear network to ensure that the derivative of the global Q-value and the individual Q-value is positive.

\begin{figure}[H]
  \includegraphics[width=10.5 cm]{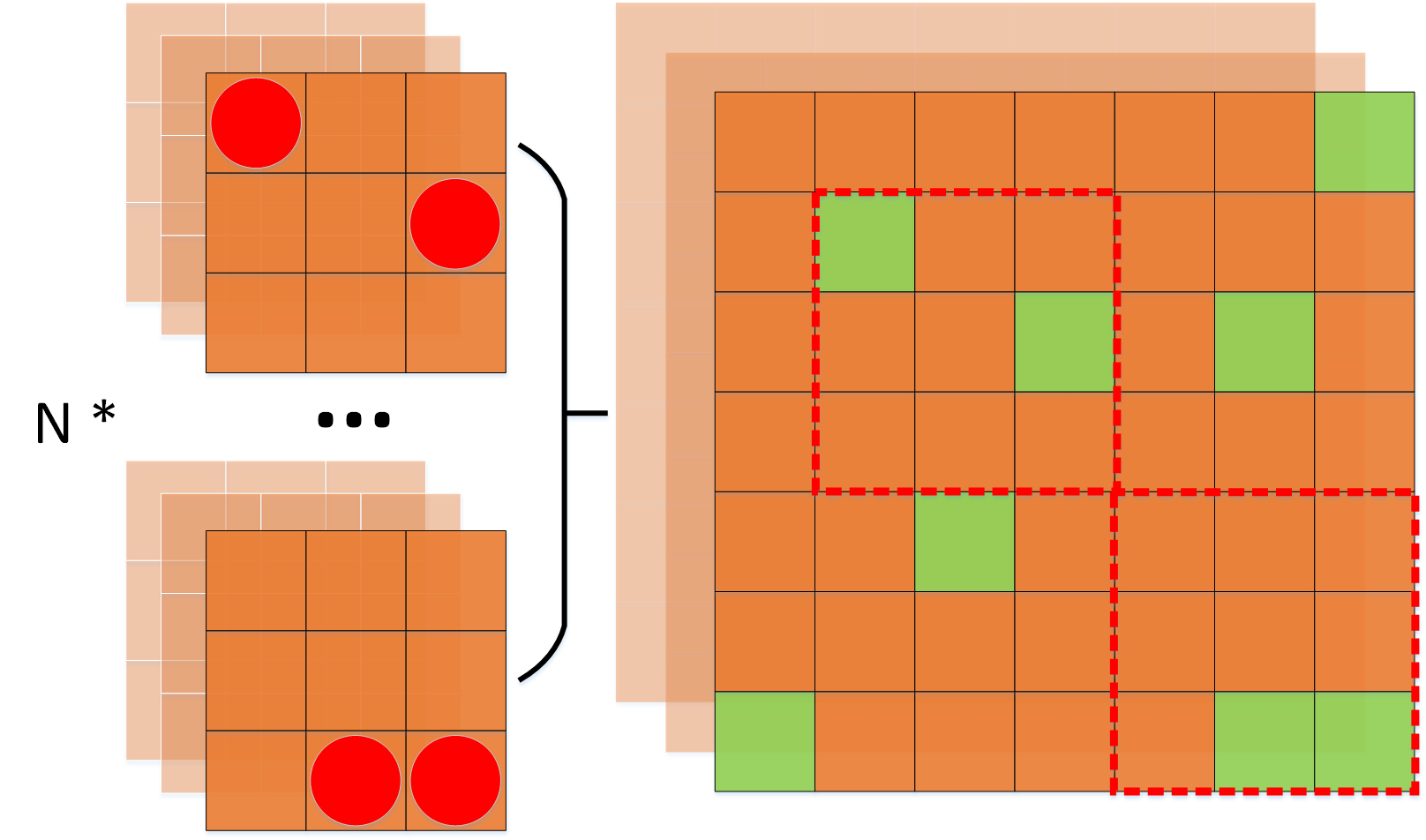}
  \caption{Generating joint observation. The observations of N pursuing vehicles are jointed into a larger observation matrix linearly.}
  \label{fig:3}
\end{figure}

\paragraph{Decision-making}

In UPDeT, in order to make the QMIX added with the transformer block more effective than the original QMIX, UPDeT adopts a policy decoupling strategy, that is, all action groups are calculated separately for the Q-value of each agent, which will make the calculation more complicated and difficult to apply to other scenarios. 

In the $ \text{T}^3 $OMVP scheme, the local information observed by each agent is jointed as shown in Figure \ref{fig:3}. After each pursuing vehicle obtains the observations of other pursuers through V2V or V2I, all the information will be jointed linearly. On the one hand, the action set in the OMVP problem is small, and it is inconvenient to divide the action set into multiple action pairs. On the other hand, the larger-range joint observation can include the set of targets observed by all pursuing vehicles to facilitate cooperative control. Therefore, $ \text{T}^3 $OMVP scheme does not have to adopt the policy decoupling strategy. The following experiments show that, in the $ \text{T}^3 $OMVP scheme, jointing the observation into the network without using policy decoupling can achieve the same benefits as adopting policy decoupling.

%%%%%%%%%%%%%%%%%%%%%%%%%%%%%%%%%%%%%%%%%%

%%%%%%%%%%%%%%%%%%%%%%%%%%%%%%%%%%%%%%%%%%
% \section{Materials and Methods}

% Materials and Methods should be described with sufficient details to allow others to replicate and build on published results. Please note that publication of your manuscript implicates that you must make all materials, data, computer code, and protocols associated with the publication available to readers. Please disclose at the submission stage any restrictions on the availability of materials or information. New methods and protocols should be described in detail while well-established methods can be briefly described and appropriately cited.

% Research manuscripts reporting large datasets that are deposited in a publicly avail-able database should specify where the data have been deposited and provide the relevant accession numbers. If the accession numbers have not yet been obtained at the time of submission, please state that they will be provided during review. They must be provided prior to publication.

% Interventionary studies involving animals or humans, and other studies require ethical approval must list the authority that provided approval and the corresponding ethical approval code.
% \begin{quote}
% This is an example of a quote.
% \end{quote}

%%%%%%%%%%%%%%%%%%%%%%%%%%%%%%%%%%%%%%%%%%
\section{Results}
\label{sec:result}
\subsection{Evader Strategy}
\label{sec:evaderstrategy}
\begin{figure} [H]
  \includegraphics[width=\linewidth]{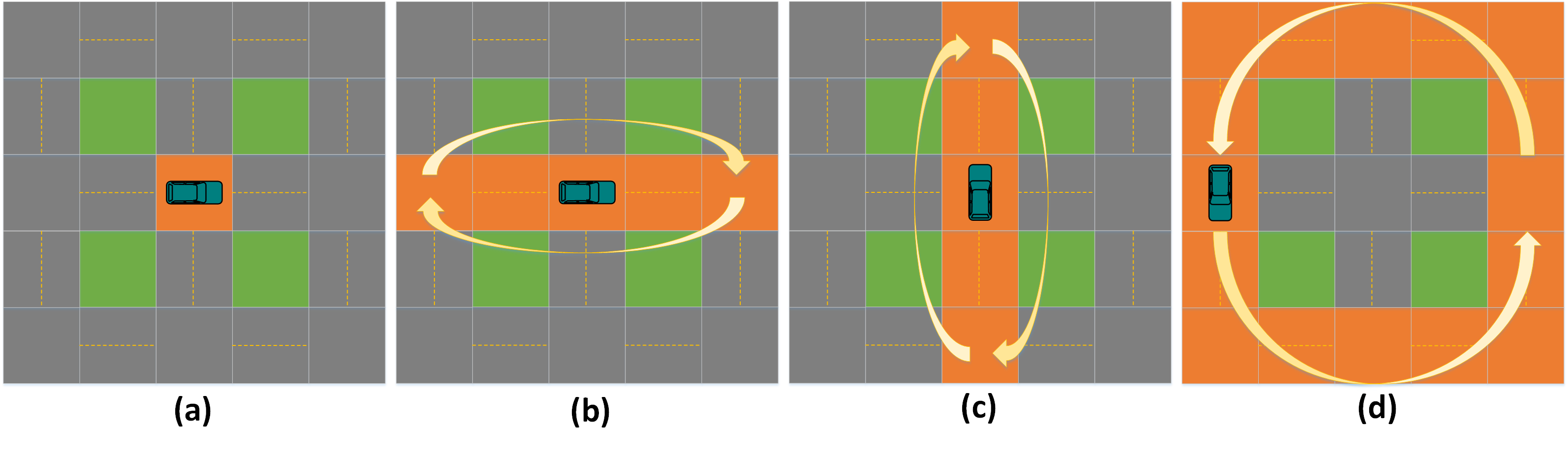}    
  \caption{Evader Strategy: (a) Keeping still. (b) Moving around in the latitudinal direction. (c) Moving around in the longitudinal direction. (d) Moving in a circle.}
  \label{fig:4}
\end{figure}

In this paper, four-movement strategies are designed for evaders to test the robustness of $ \text{T}^3 $OMVP in the pursuit. The four strategies are: keeping still, moving around in the latitudinal direction, moving around in the longitudinal direction, and moving in a circle as shown in Figure \ref{fig:4}. In each episode of training and testing, the evading vehicle randomly selects and executes one strategy from the above four.

\subsection{$ \epsilon $-greedy}
$ \text{T}^3 $OMVP utilizes the $ \epsilon $-greedy strategy when the agent makes a decision. Specifically, the agent randomly chooses an unknown action with probability $ \epsilon $ to explore or exploits with probability 1-$ \epsilon $ by selecting the action with the largest Q-value among the existing actions. The $ \epsilon $-greedy algorithm is as follows:
\begin{equation}\label{eqexpmuts:9}
\begin{aligned}
A^{*} & = \arg \max _{a} Q(s, a) \\
\pi(a \mid s) & = \begin{cases}1-\epsilon+\epsilon /|\mathcal{A}(s)| & \text { if } a=A^{*} \\ \epsilon /|\mathcal{A}(s)| & \text { if } a \neq A^{*}\end{cases}.
\end{aligned}
\end{equation}
Where $ A^{*} $ represents the local optimal action obtained according to the Q-value.

\subsection{Simulation Settings}
The $ \text{T}^3 $OMVP scheme is trained and evaluated in a self-designed urban multi-intersection environment, developed with Python 3.6. The environment is based on the gym and contains multiple intersections in a $ W * W $ grid environment. In the experiment, $ \lambda $ represents the ratio of pursuing vehicles to evading vehicles. In order to verify the effect of $ \lambda $ on the stability of $ \text{T}^3 $OMVP scheme, $ \lambda $ is set variably with 2, 1, 0.5. Experimental scenarios include the 8v4 (eight pursuing vehicles vs four evading vehicles) scenario, 4v4 (four pursuing vehicles vs four evading vehicles) scenario, and 2v4 (two pursuing vehicles vs four evading vehicles) scenario. The parameters of the relevant experimental environment are shown in \mbox{Table \ref{tab:Experiment_Parameter}}. 
% All source codes about the proposed $ \text{T}^3 $OMVP scheme are provided on GitHub (\url{https://github.com/pipihaiziguai/T3OMVP}). 

\begin{table} [H]
% \caption{Experimental Parameters\label{tab1}}
\caption{Experimental Parameters.}
\label{tab:Experiment_Parameter}
\newcolumntype{C}{>{\arraybackslash}X}
\begin{tabularx}{\textwidth}{ l  l }
\toprule
\textbf{Parameters}	& \textbf{Value}\\
\midrule
Time step         & $ 50 $    \\
Grid space width $ W $       & $ [13,17,21] $    \\
Intersection interval        & $ 1 $     \\
The distance of vehicle moves each time    & $ 1 $     \\
The size of the observation space of the pursuing vehicle  & $ 5*5 $   \\
The size of the joint observation space of the pursuing vehicle & $ [3*13*13, 3*17*17, 3*21*21] $ \\
Number of evading vehicles       & $ 4 $ \\
Historical observation length    & $ 5 $ \\
\bottomrule
\end{tabularx}
\end{table}
\unskip
This paper adopts QMIX as the baseline method, including two hypernetworks, and uses the $ Elu $ activation function after each layer of the network $ DN(o) $ where $ o $ represents the dimension of the network output. The $ Elu $ can avoid the disappearance of the gradient, reduce the training time, and improve the accuracy in the neural network \cite{clevert2015fast}. Similar to UPDeT, after obtaining the joint observation, $ \text{T}^3 $OMVP feeds it into the transformer together with the hidden state. The transformer encoder consists of multiple blocks, each of which contains a multi-head self-attention, a feed-forward neural network, and two resnet structures. In UPDeT, QMIX with transformer needs to adopt the policy decoupling strategy to achieve better performance than QMIX with transformer. However, $ \text{T}^3 $OMVP adopts an extensive observation matrix integrating observations of all pursuing vehicles to replace policy decoupling strategy, and achieves a better or same performance than UPDeT. The complete hyperparameters are listed in Table \ref{HyperParameter}.

\begin{table} [H]
% \caption{Networks Parameters\label{tab2}}
\caption{Hyperparameters for Neural Networks.}
\label{HyperParameter}
\newcolumntype{C}{>{\arraybackslash}X}
\begin{tabularx}{\textwidth}{ l  l }
% \begin{tabularx}{\textwidth}{CCC}
\toprule
\textbf{Hyperparameters}	& \textbf{Value}\\
\midrule
Batch size             & $ 32 $                            \\
Learning rate          & $ 0.001 \rightarrow 0 $       \\
Optimizer              & Adam                          \\
Discounted factor $ \gamma $    & $ 0.95 $                          \\
$ \epsilon $ decay          & $ 0.0001 $                        \\
$ \epsilon $ min            & $ 0.1 $                           \\
\midrule
\textbf{\emph{QMIX}} \\
Hypernetwork w \#1     & $ {[}DN(128), Elu, DN(128+N){]} $ \\
Hypernetwork w \#final & $ {[}DN(128), Elu, DN(128){]} $   \\
Hypernetwork b \#1     & $ {[}DN(128){]} $                 \\
Output                 & $ {[}DN(128), Elu, DN(1){]} $     \\
\midrule
\textbf{\emph{Transformer Encoder}} \\
Transformer depth         & $ 2 $                          \\
Embedding vector length                  & $ 250 $         \\
Number of heads                & $ 5 $                      \\      
\bottomrule
\end{tabularx}
\end{table}
\unskip

\subsection{Discussion}
\label{sec:discussion}
The entire experiment is divided into three parts. In the first part, to compare $ \text{T}^3 $OMVP with UPDeT, the unified reward function is used to evaluate $ \text{T}^3 $OMVP, QMIX, QMIX+UPDeT, VDN, VDN+UPDeT, and $ \text{T}^3 $VDN except M3DDPG which has been proven not to perform well in the pursuit-evasion scenario \cite{resilient}.
In this part, all methods are trained in the $13*13$ multi-vehicle pursuit grid environment with 8v4 scenario, 4v4 scenario, and 2v4 scenario, respectively. There are four strategies for evading vehicles as shown in Section \ref{sec:evaderstrategy}. 
At test time, 50 episodes of verification are performed to calculate the average value of the reward as the reward for the current training step.
In the second part, the performances of $ \text{T}^3 $OMVP, QMIX, and VDN are verified on multiple scenarios of different difficulty, including $ \lambda \in \{2,1,0,5\} $ and $ W \in \{13,17,21\} $. 
Finally, to evaluate the role of self-attention mechanism in decision-making, the attention is analyzed in two aspects: team observations and time observations.
 
\begin{figure}
  \includegraphics[width=\linewidth]{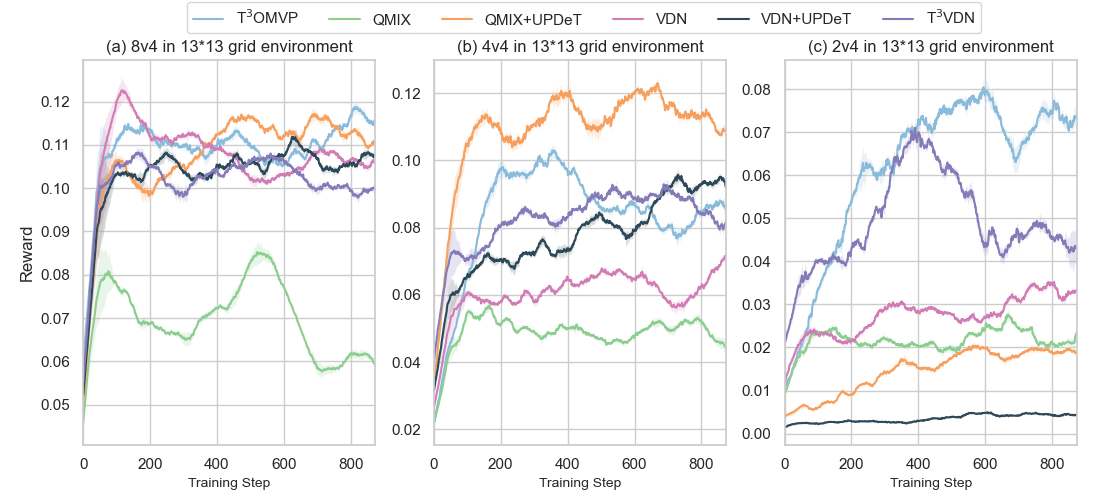} 
  \caption{Comparison of $ \text{T}^3 $OMVP and UPDeT methods in the $13*13$ grid environment on 8v4 scenario, 4v4 scenario, and 2v4 scenario, respectively. QMIX+UPDeT and VDN+UPDeT refer to methods using the UPDeT with policy decoupling strategy. $ \text{T}^3 $VDN refers to VDN using transformer-based time and team observations without policy decoupling strategy.}
  \label{fig:5}
\end{figure}

Figure \ref{fig:5} depicts performance comparison of $ \text{T}^3 $OMVP, QMIX, QMIX+UPDeT, VDN, VDN+UPDeT, and $ \text{T}^3 $VDN in the $ 13*13 $ grid environment on 8v4 scenario, 4v4 scenario, and 2v4 scenario, respectively. 
As shown in Figure \ref{fig:5} (a-c), the $ \text{T}^3 $OMVP scheme, based on QMIX and using joint observation rather than policy decoupling strategy, can achieve the best performance on both the simple 8v4 scenario and the difficult 2v4 scenario. However, its performance on the 4v4 scenario is not as good as that of QMIX+UPDeT which uses the policy decoupling strategy. The reason is that on 4v4 scenario, the observations dimension is close to the number of actions, so the action-observation pairs are more suitable for policy decoupling. However, when the observations dimension and the number of actions are different, policy decoupling will make it difficult to assign appropriate actions, resulting in performance degradation even worse than original QMIX on 2v4 scenario. 
Furthermore, $ \text{T}^3 $OMVP weakens the influence of observations dimension and number of actions, since the transformer combined with $ TT-observations $ endows the pursuing vehicle with more comprehensive information from all other vehicles, which avoids vehicles from falling into the local optimum situation, thereby, the pursuing efficiency can be improved in more scenarios.
The purple line and dark blue line in Figure \ref{fig:5} indicate that VDN+UPDeT performs poorly on the OMVP scenario. This phenomenon can be interpreted as follows: on the one hand, the relatively simple reward mechanism in our scenario is more adaptive with a concise way of calculating the global Q-value, thence, VDN can solely achieve relatively competitive performance. On the other hand, the update process involved in the VDN is relatively simple, thus VDN+UPDeT is not sufficient for significantly improving the original performance on the OMVP scenario. Given the result that VDN combined with a transformer structure and $TT-observations$ can still achieve better performance than VDN, it proves that transformer combined with $TT-observations$ can improve or maintain the performance without policy decoupling strategy.

\begin{figure}
  \includegraphics[width=\linewidth]{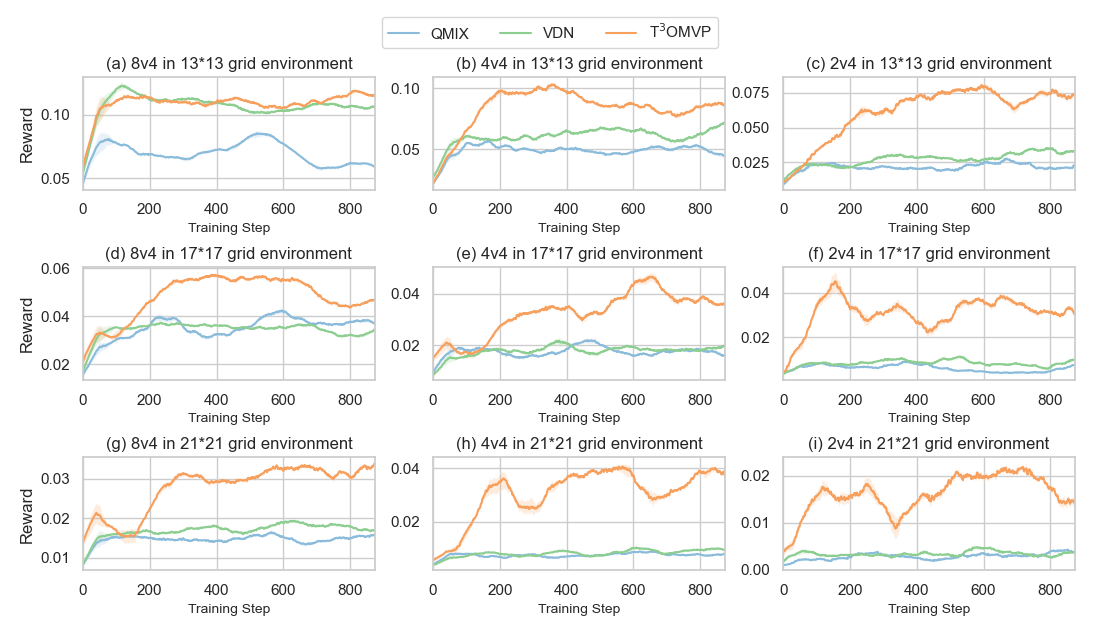} 
  \caption{Comparison of $ \text{T}^3 $OMVP, QMIX, and VDN on 8v4 scenario, 4v4 scenario, and 2v4 scenario with $ W \in \{13,17,21\} $.}
  \label{fig:6}
\end{figure}

Figure \ref{fig:6} shows the performances of $ \text{T}^3 $OMVP, QMIX, and VDN at different difficulty levels. According to the change of $ \lambda $, Figure \ref{fig:6} (a) (b) (c) show the situation in a $13*13$ grid environment on 8v4 scenario, 4v4 scenario, and 2v4 scenario, respectively.
% Figure \ref{fig:6} (a) shows the situation of eight pursuing vehicles and four evading vehicles in a $13*13$ grid environment, and Figure \ref{fig:6} (b) shows the situation of four pursuing vehicles and four evading vehicles in a $13*13$ grid environment, Figure \ref{fig:6} (c) shows the situation of two chasing vehicles and four evading vehicles in the $13*13$ grid environment.
As shown in  Figure \ref{fig:6} (a-c), on the simple 8v4 scenario, VDN shows better performances than QMIX and can even reach the performance of $ \text{T}^3 $OMVP. It indicates that VDN can adapt to such situations better than QMIX in relatively simple scenarios.
As the number of pursuing vehicles decreases, the problem becomes complicated. The performance gap between $ \text{T}^3 $OMVP and QMIX gradually widens, and so as that between $ \text{T}^3 $OMVP and VDN. Furthermore, the performance gap between VDN and QMIX gradually narrows. The results indicate that as the difficulty increases, VDN with a simple structure cannot adapt to these scenarios anymore, thus the gap betweent it and QMIX gradually narrows, but $ \text{T}^3 $OMVP can still achieve the best performance. According to the change of $ W $, Figure \ref{fig:6} (d) (g) show the situation on 8v4 scenario in the $17*17$ grid environment and $21*21$ grid environment, respectively. As shown in Figure \ref{fig:6} (a) (d) (g), it can be seen that as the size of the environment increases, the difficulty of pursuing increases, VDN still gradually reaches the same performance as QMIX, and $ \text{T}^3 $OMVP gradually shows better performance than QMIX and VDN. It is consistent with the previous statement that $ \text{T}^3 $OMVP can show better performance in more difficult scenarios, since $ \text{T}^3 $OMVP can more comprehensively evaluate the observed information through the self-attention mechanism. At the same time, as the difficulty of the scenario increases, the gap between QMIX and VDN gradually narrows, and QMIX shows more competitive performance than VDN on 2v4 scenario in the 21*21 grid environment. It can be concluded that, as the difficulty increases, QMIX can adapt to more complex scenarios, which is beneficial to the decision-making among multi-vehicles. 
Furthermore, this paper compaperd the performance of $ \text{T}^3 $OMVP and QMIX on multiple test scenarios. In order to eliminate the influence of the evading vehicle movement, the strategy of the evading vehicles is keeping still and the random number seed is fixed. The results are shown in Table \ref{tab:3}. It is observed that transformer structure improves the performance of QMIX by 9.66\%\textasciitilde106.25\% on multiple test scenarios.
For the reason that, $ \text{T}^3 $OMVP scheme is generated by adding the transformer structure on the basis of QMIX to deal with the OMVP problem in the urban multi-intersection environment. 

\begin{table} [H]
\caption{Testing for comparison of performance on 8v4 scenario, 4v4 scenario, and 2v4 scenario with $ W \in \{13,17,21\} $.}
\label{tab:3}
\centering
\begin{tabular}{lllll}
\hline
\multicolumn{1}{|l|}{Sum of rewards}         &\multicolumn{1}{l|}{Grid space width} & \multicolumn{1}{l|}{13}      & \multicolumn{1}{l|}{17}       & \multicolumn{1}{l|}{21}      \\ \hline
\multicolumn{1}{|l|}{\multirow{3}{*}{8v4 scenario}} & \multicolumn{1}{l|}{QMIX}        & \multicolumn{1}{l|}{3.88}    & \multicolumn{1}{l|}{1.1775}   & \multicolumn{1}{l|}{0.76}    \\ \cline{2-5} 
\multicolumn{1}{|l|}{}                              & \multicolumn{1}{l|}{$ \text{T}^3 $OMVP}          & \multicolumn{1}{l|}{4.255}   & \multicolumn{1}{l|}{1.5175}   & \multicolumn{1}{l|}{0.9175}  \\ \cline{2-5} 
\multicolumn{1}{|l|}{}                              & \multicolumn{1}{l|}{Improvement} & \multicolumn{1}{l|}{9.66\%}  & \multicolumn{1}{l|}{28.87\%}  & \multicolumn{1}{l|}{20.72\%} \\ \hline
\multicolumn{1}{|l|}{\multirow{3}{*}{4v4 scenario}} & \multicolumn{1}{l|}{QMIX}        & \multicolumn{1}{l|}{1.345}   & \multicolumn{1}{l|}{0.455}    & \multicolumn{1}{l|}{0.345}   \\ \cline{2-5} 
\multicolumn{1}{|l|}{}                              & \multicolumn{1}{l|}{$ \text{T}^3 $OMVP}          & \multicolumn{1}{l|}{2.115}   & \multicolumn{1}{l|}{0.88}     & \multicolumn{1}{l|}{0.475}   \\ \cline{2-5} 
\multicolumn{1}{|l|}{}                              & \multicolumn{1}{l|}{Improvement} & \multicolumn{1}{l|}{57.24\%}   & \multicolumn{1}{l|}{93.40\%}  & \multicolumn{1}{l|}{37.68\%} \\ \hline
\multicolumn{1}{|l|}{\multirow{3}{*}{2v4 scenario}} & \multicolumn{1}{l|}{QMIX}        & \multicolumn{1}{l|}{0.51}    & \multicolumn{1}{l|}{0.16}     & \multicolumn{1}{l|}{0.09}    \\ \cline{2-5} 
\multicolumn{1}{|l|}{}                              & \multicolumn{1}{l|}{$ \text{T}^3 $OMVP}          & \multicolumn{1}{l|}{0.86}    & \multicolumn{1}{l|}{0.33}     & \multicolumn{1}{l|}{0.18}    \\ \cline{2-5} 
\multicolumn{1}{|l|}{}                              & \multicolumn{1}{l|}{Improvement} & \multicolumn{1}{l|}{68.62\%} & \multicolumn{1}{l|}{106.25\%} & \multicolumn{1}{l|}{100\%}   \\ \hline
\end{tabular}
\end{table}

\begin{figure} [H]
  \includegraphics[width=10.5 cm]{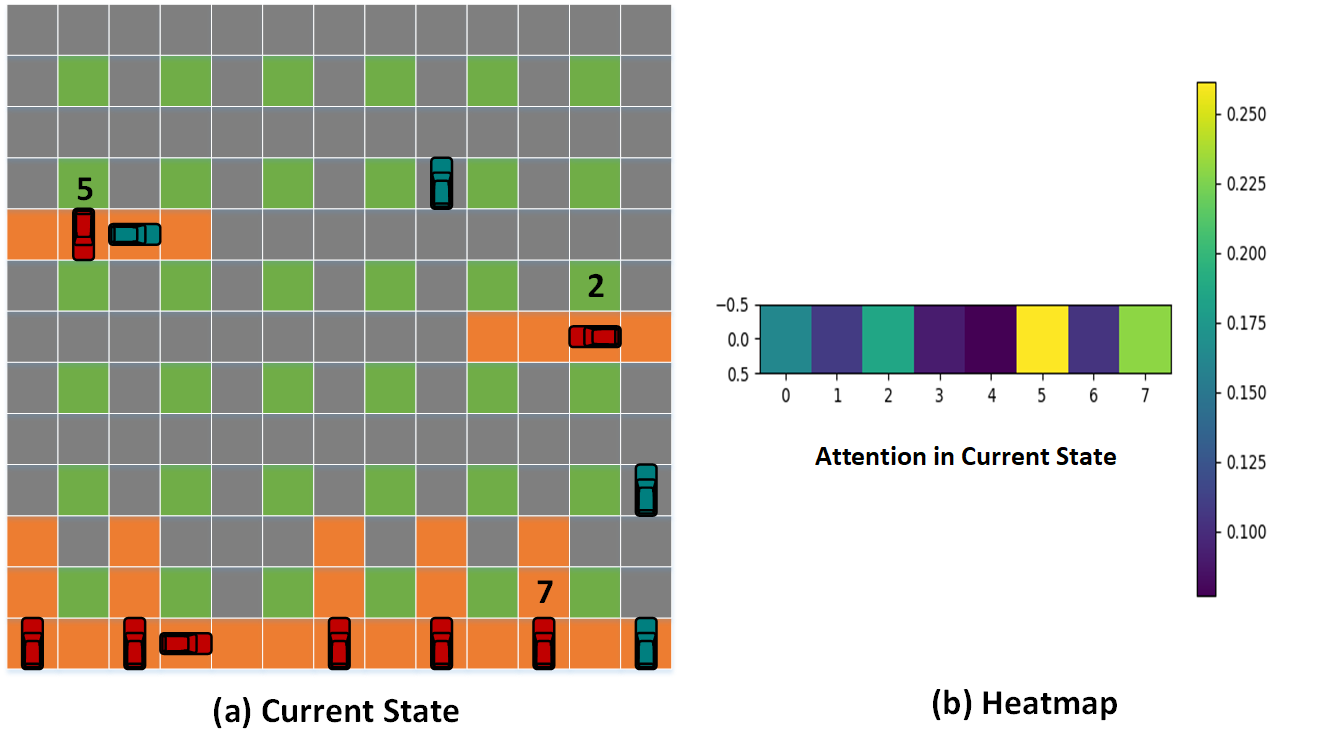}    
  \caption{Heatmap of Team Attention. (a) shows the current state in which the red agent is the pursuing vehicle, the blue agent is the evading vehicle, the gray area is the road, the yellow area is the observation of the pursuing vehicle, and the green area is the building. (b) shows the attention in the current state.}
  \label{fig:7}
\end{figure}

Figure \ref{fig:7} (a) describes the 8v4 scenario in the 13*13 urban multi-intersection environment. As shown in Figure \ref{fig:7} (b), when two pursuing vehicles observe different targets, the attention of the pursuing vehicle No. 5 is the highest and that of the pursuing vehicle No.7 ranks second. The reason is pursuing vehicles No. 5 and No. 7 are the closest vehicles to the evading vehicle and the calculated attention is also biased towards No. 5 and No. 7. In addition, Figure \ref{fig:7} (b) also shows that the attention of the pursuing vehicle No. 2 is the highest except for the pursuing vehicles No. 5 and No. 7. The reason is that pursuing vehicle No. 2 is most relevant to the direction of the evading vehicle observed by No. 7. Therefore, focusing on pursuing vehicle No. 2 can improve the pursuing efficiency by mobilizing the No. 2 vehicle to pursue the evading vehicle observed by No. 7 in the subsequent actions. 

\begin{figure} [H]
  \includegraphics[width=\linewidth]{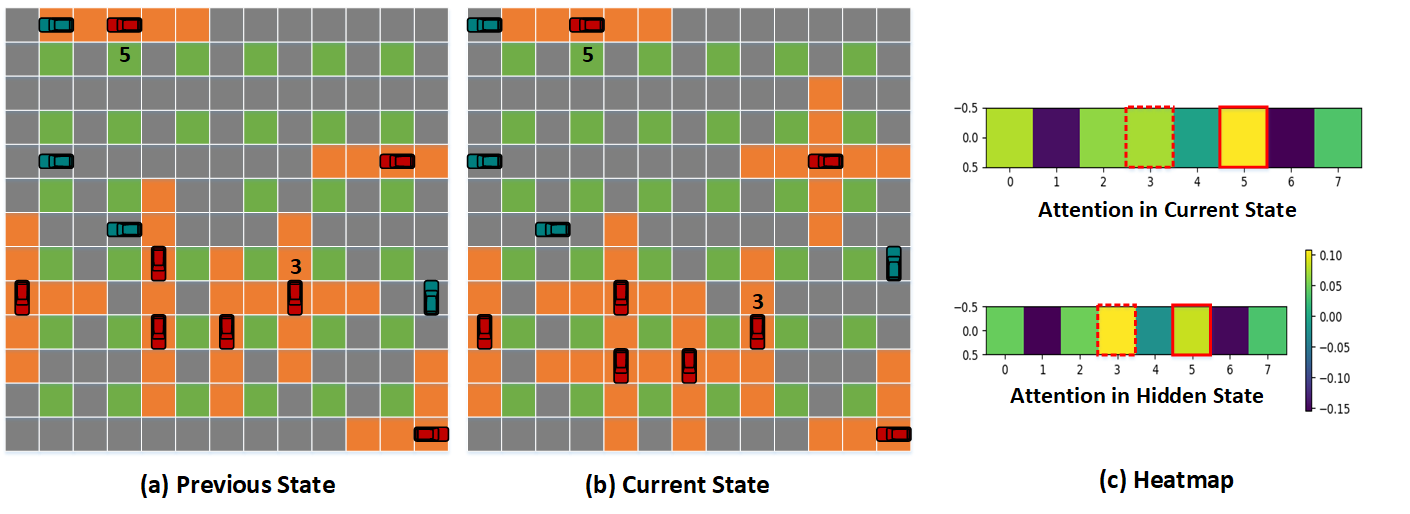}    
  \caption{Heatmap of Hidden State. (a) (b) show the previous state and current state. The upper of (c) shows the attention in current state. The bottom of (c) shows the attention in hidden state of previous state.}
  \label{fig:8}
\end{figure}

Figure \ref{fig:8} shows the role of hidden states in historical observations. Figure \ref{fig:8} (a) (b) show the positional relationship between the pursuing vehicle and the evading vehicle in the previous state and current state respectively, and (c) shows the attention in the current state and the hidden state retained in the previous state. As shown in Figure \ref{fig:8} (a) (b), since the pursuing vehicle No. 5 did not move, the evading vehicle observed by the pursuing vehicle No. 5 in the previous step drove out of the observation field of the pursuing vehicle No. 5, however, the pursuing vehicle No. 5 still has the highest attention among the attentions in the current state. Due to the hidden state of previous state, the attention reserved by the pursuing vehicle No. 5 is the highest except for the pursuing vehicle No. 3, since the pursuing vehicle No. 3 does not observe the evading vehicle in both steps and the observation field shrinks due to the complicated road structure, that is why the pursuing vehicle No. 5 can get the highest attention at the current state. According to the transmission of the hidden state, the pursuing vehicle can save the experience of historical observations and improve the pursuing efficiency.

%%%%%%%%%%%%%%%%%%%%%%%%%%%%%%%%%%%%%%%%%%
% conclusions
\section{Conclusions}
\label{sec:conclusion}

An OMVP problem is defined from MVP problem in which the pursuing vehicles can communicate freely and are required to pursue the evading vehicles quickly, whereas the observation is blocked by urban buildings.
A $ \text{T}^3 $OMVP scheme is proposed to solve the OMVP problem in the urban multi-intersection environment. The $ \text{T}^3 $OMVP scheme uses QMIX as the baseline and utilizes the transformer structure to process the time and team's attention to various observations. Different from UPDeT, $ \text{T}^3 $OMVP uses joint observation to collect the observations of all pursuing vehicles instead of the policy decoupling strategy for final attention. $\text{T}^3 $OMVP can directly make decisions and achieve a better or same performance without policy decoupling. This paper modifies the predator-prey scenario and obtains a complex urban multi-intersection OMVP simulator, which is a light-weight system and easy to train agents, and the proposed $\text{T}^3 $OMVP scheme is verified in this simulation environment. More extensive experiments prove that $ \text{T}^3 $OMVP can achieve much better performance on more difficult scenarios than original QMIX and VDN.

The future work will focus on the modeling and theoretical analysis of transformers and MARL. It will greatly help to better explain the application of the attention mechanism in processing joint observation among agents in MARL.

%%%%%%%%%%%%%%%%%%%%%%%%%%%%%%%%%%%%%%%%%%
\authorcontributions{Conceptualization, Z.Y. and L.Z.; methodology, Z.Y. and T.W.; software, Z.Y.; validation, Z.Y., Q.W. and Y.Y.; investigation, Z.Y., Q.W. and Y.Y.; resources, L.Z.; data curation, Z.Y., Q.W. and Y.Y.; writing---original draft preparation, Z.Y.; writing---review and editing, Z.Y., Q.W., Y.Y. and L.L.; visualization, Z.Y.; supervision, L.Z. and L.L.; project administration, Z.Y. All authors have read and agreed to the published version of the manuscript.}

\funding{This work was partially supported by the National Natural Science Foundation of China (Grant No. 62176024).}

\dataavailability{The code is open-sourced at \url{https://github.com/pipihaiziguai/T3OMVP}.} 

% \acknowledgments{In this section you can acknowledge any support given which is not covered by the author contribution or funding sections. This may include administrative and technical support, or donations in kind (e.g., materials used for experiments).}

\conflictsofinterest{The authors declare no conflict of interest.} 

%% Optional
% \sampleavailability{Samples of the compounds ... are available from the authors.}

%%%%%%%%%%%%%%%%%%%%%%%%%%%%%%%%%%%%%%%%%%
%% Only for journal Encyclopedia
%\entrylink{The Link to this entry published on the encyclopedia platform.}

%%%%%%%%%%%%%%%%%%%%%%%%%%%%%%%%%%%%%%%%%%
%% Optional
\abbreviations{Abbreviations}{
The following abbreviations are used in this manuscript:\\

\noindent 
\begin{tabular}{@{}ll}
IoVs & Smart Internet of Vehicles\\
MDP & Markov Decision Processes\\
MVP & Multi-Vehicle Pursuit\\
MARL & Multi-agent Reinforcement Learning \\
CTDE & Centralized Training with Decentralized Execution \\
OMVP & Observation-constrained Multi-Vehicle Pursuit\\
$ \text{T}^3 $OMVP & Transformer-based Time and Team Reinforcement Learning Scheme for OMVP \\
Dec-POMDP & Decentralized Partially Observed Markov Decision Processes

\end{tabular}}

\begin{adjustwidth}{-\extralength}{0cm}

\reftitle{References}

\end{adjustwidth}
\end{document}